\documentclass{article}

\usepackage[nonatbib,final]{neurips_2019}

\usepackage[utf8]{inputenc} 
\usepackage[T1]{fontenc}    
\usepackage{hyperref}       
\usepackage{url}            
\usepackage{booktabs}       
\usepackage{amsfonts}       
\usepackage{nicefrac}       
\usepackage{microtype}      
\usepackage{amsmath,amsthm,amssymb}
\usepackage{caption} 
\usepackage{bookmark}
\usepackage{graphicx}
\usepackage{float}
\usepackage{amsfonts}
\usepackage{xcolor}
\usepackage{caption}
\usepackage{subcaption}
\usepackage{cleveref}
\usepackage{appendix}

\usepackage[utf8]{inputenc}
\usepackage[english]{babel}

\newtheorem{theorem}{Theorem}[section]
\newtheorem{corollary}{Corollary}[theorem]

\newtheorem{defn}[theorem]{Definition}
\newtheorem{prop}[theorem]{Proposition}

\newtheorem*{remark}{Remark}

\usepackage{mathtools}
\DeclarePairedDelimiter{\ceil}{\lceil}{\rceil}

\title{Statistical Analysis of Nearest Neighbor Methods for Anomaly Detection}

\author{%
  Xiaoyi Gu \\
  Department of Statistics\\
  Carnegie Mellon Univeristy\\
  Pittsburgh, PA 15213 \\
  \texttt{xgu1@andrew.cmu.edu} \\
   \And
  Leman Akoglu \\
  Heinz College of Information Systems and Public Policy\\
  Carnegie Mellon Univeristy\\
  Pittsburgh, PA 15213 \\
  \texttt{lakoglu@andrew.cmu.edu} \\
   \AND
  Alessandro Rinaldo \\
  Department of Statistics\\
  Carnegie Mellon Univeristy\\
  Pittsburgh, PA 15213 \\
  \texttt{arinaldo@cmu.edu } \\
}

\begin{document}

\maketitle

\begin{abstract}
Nearest-neighbor (NN) procedures are well studied and widely used in both supervised and unsupervised learning problems. In this paper we are concerned with investigating the performance of NN-based methods for anomaly detection. We first show through extensive simulations that NN methods compare favorably to some of the other state-of-the-art algorithms for anomaly detection based on a set of benchmark synthetic datasets. We further consider the performance of NN methods on real datasets, and relate it to the dimensionality of the problem. Next, we analyze the theoretical properties of NN-methods for anomaly detection by studying a more general quantity called distance-to-measure (DTM), originally developed in the literature on robust geometric and topological inference. We provide finite-sample uniform guarantees for the empirical DTM and use them to derive misclassification rates for anomalous observations under various settings. In our analysis we rely on Huber's contamination model and formulate mild geometric regularity assumptions on the underlying distribution of the data.
\end{abstract}

\section{Introduction}
Anomaly detection is the process of detecting instances that deviate significantly from the other sample members. The problem of detecting anomalies can arise in many different applications, such as fraud detection in financial transactions, intrusion detection for security systems, and various medical examinations. 

Depending on the availability of data labels, there are multiple setups for anomaly detection. The first is the supervised setup, where labels are available for both normal and anomalous instances during the training stage. Because of its similarity to the standard classification setup, numerous classification methods with good empirical performance and well-studied theoretical properties can be adopted. The second setup is the semi-supervised setup, where training data only comprise normal instances and no anomalies. Well-known methods with theoretical guarantees include $k$NNG \cite{GEM}, BP-$k$NNG \cite{Efficient_GEM} and BCOPS \cite{conformal}, with the first two methods developed based on the geometric entropy minimization (GEM) principle proposed in \cite{GEM}, and the third on conformal prediction. The third setup is the unsupervised setup, which is the most flexible yet challenging setup. For the rest of the paper, we will only focus on this setup and do not assume any prior knowledge on data labels.

Many empirical methods have been developed in the unsupervised setup, which can be roughly classified into four categories: density based methods such as the Robust KDE ($\mathrm{RKDE}$) \cite{RKDE}, Local Outlier Factor ($\mathrm{LOF}$) \cite{LOF}, and mixture models ($\mathrm{EGMM}$); distance based methods such as $k\mathrm{NN}$ \cite{knn} and Angle-based Outlier Detection ($\mathrm{ABOD}$) \cite{abod}; model based methods such as the one-class SVM ($\mathrm{OCSVM}$) \cite{ocsvm}, $\mathrm{SVDD}$ \cite{svdd}, and autoencoders \cite{autoencoder}; ensemble methods such as Isolation Forest ($\mathrm{IForest}$) \cite{IF} and $\mathrm{LODA}$ \cite{LODA}. In practice, ensemble methods are often times favored for their computational efficiency and robustness to tuning parameters, yet there is little theoretical understanding of how and why these algorithms work. 

In this paper, we focus on studying NN-methods in the unsupervised setting. We begin with an empirical analysis of NN-methods on a set of synthetic benchmark datasets and show that they compare favorably to the other state-of-the-art algorithms. We further discuss their performance on real datasets and relate it to the dimensionality of the problem. Next, we provide statistical analysis of NN-methods by analyzing the distance-to-a-measure (DTM) \cite{dtm_first}, a generalization to the NN scheme. The quantity was initially raised in the robust topological inference literature, in which DTM proves to be an effective distance-like function for shape reconstruction in the presence of outliers \cite{dtm_paper}. We give finite sample uniform guarantees on the empirical DTM, and also demonstrate how DTM classifies the anomalies, under suitable assumptions on the underlying distribution of the data.

\section{Empirical Performance of NN-methods}

Two versions of the NN anomaly detection algorithms have been proposed: $k^{\mathrm{th}}\mathrm{NN}$ \cite{kthNN} and $k\mathrm{NN}$ \cite{knn}. $k^{\mathrm{th}}\mathrm{NN}$ assigns anomaly score of an instance by computing the distance to its $k^{\text{th}}$-nearest-neighbor, whereas $k\mathrm{NN}$ takes the average distance over all $k$-nearest-neighbors. Both methods are shown to have competitive performance in various comparative studies \cite{compare, compare2, LODA, compare3}. In particular, the comparative study developed by Goldstein and Uchida \cite{compare} is the one of most comprehensive analysis to date that includes the discussion of NN-methods and, at the same time, aligns with the unsupervised anomaly detection setup. However, the authors omit the analysis of ensemble methods, some of which are considered as state-of-the-art algorithms (e.g., $\mathrm{IForest}$ and $\mathrm{LODA}$). Emmott et al. \cite{meta} constructed a large corpus (over 20,000) of synthetic benchmark datasets that vary across multiple aspects (e.g., clusteredness, separability, difficulty, etc). The authors evaluate the performance of eight top-performing algorithms, including $\mathrm{IForest}$ and $\mathrm{LODA}$, but omit the analysis of NN-methods. In this section, we provide a comprehensive empirical analysis of NN-methods by comparing $k\mathrm{NN}$, $k^{\mathrm{th}}\mathrm{NN}$, and $\mathrm{DTM}_2$\footnote{$\mathrm{DTM}_2$ stands for the empirical DTM (see Section \ref{sec:theory}) with $q=2$. We include its empirical analysis here for comparison purposes.} to $\mathrm{IForest}$, $\mathrm{LOF}$ and $\mathrm{LODA}$ on (1) the corpus of synthetic datasets developed in \cite{meta}, (2) 23 real datasets from the ODDS library \cite{ODDS}, and (3) 6 high dimensional datasets from the UCI library \cite{UCI}. The code for all our experiments will be made publicly available. In general, no one methodology should be expected to performs well in all possible scenarios. In Section 4 of the Appendix we present different examples in which IForest, LODA, LOF and $\mathrm{DTM}_2$ perform very differently.
\subsection{Comparison on Benchmark Datasets} \label{sec:sim}
First, we complement Emmott et al.’s study \cite{meta} by extending it to NN-based detectors. First, we calculate the ROC-AUC (AUC) and Average Precision (AP) scores for each method on each benchmark, and compute their respective quantiles on the empirical distributions for AUC and AP scores (refer to Appendix E in \cite{meta} for more details on treating AUC and AP as random variables). We say that an algorithm fails on a benchmark with metric AUC (or AP) at significance level $\alpha$ if the computed AUC (or AP) quantiles are less than $(1-\alpha)$. Then, the failure rate for each algorithm is found as the percentage of failures over the entire benchmark corpus. The results are shown in Table \ref{benchmark_table}, where the top section is copied from \cite{meta} and the bottom section shows the failure rates we obtained for $k\mathrm{NN}$, $k^{\mathrm{th}}\mathrm{NN}$, and $\mathrm{DTM}_2$. The "Either" column indicates that the benchmarks fail under at least one of the two metrics. For all three methods, $k$ is set as $0.03\times($sample size), same as the parameter the authors used for $\mathrm{LOF}$. Among all methods, $\mathrm{IForest}$ gives the lowest failure rates (boldfaced) for all three metrics. $k\mathrm{NN}$ and $\mathrm{DTM}_2$ turn out to be next-best top performers, falling marginally behind $\mathrm{IForest}$.
\begin{table}
\centering
\caption{Algorithm Failure Rate with Significance Level $\alpha = 0.001$.}
\begin{tabular}{|c||c|c|c|}
\hline
            & \textbf{AUC}    & \textbf{AP}     &\textbf{Either}  \\ \hline \hline
$\mathrm{ABOD}$        & 0.5898 & 0.6784 & 0.7000 \\ 
$\mathrm{IForest}$     & \textbf{0.5520} & \textbf{0.6514} & \textbf{0.6741} \\ 
$\mathrm{LODA}$        & 0.6187 & 0.6955 & 0.7194 \\
$\mathrm{LOF}$         & 0.6016 & 0.7071 & 0.7331 \\ 
$\mathrm{RKDE}$        & 0.6122 & 0.7030 & 0.7194 \\ 
$\mathrm{OCSVM}$       & 0.7218 & 0.7342 & 0.7969 \\ 
$\mathrm{SVDD}$        & 0.8482 & 0.8868 & 0.9080 \\ 
$\mathrm{EGMM}$        & 0.6188 & 0.7146 & 0.7303 \\ \hline \hline
$k\mathrm{NN}$      & 0.5646 & 0.6744 & 0.6960 \\ 
$k^{\mathrm{th}}\mathrm{NN}$ & 0.5831 & 0.6886 & 0.7100\\
$\mathrm{DTM}_2$ & 0.5669 & 0.6761 & 0.6977\\
\hline
\end{tabular}
\label{benchmark_table}
\end{table}

\subsection{Comparison on Datasets from the ODDS library} \label{sec:real}
\begin{figure}
    \centering
    \begin{subfigure}[t]{0.49\textwidth}
        \centering
        \includegraphics[width=\textwidth]{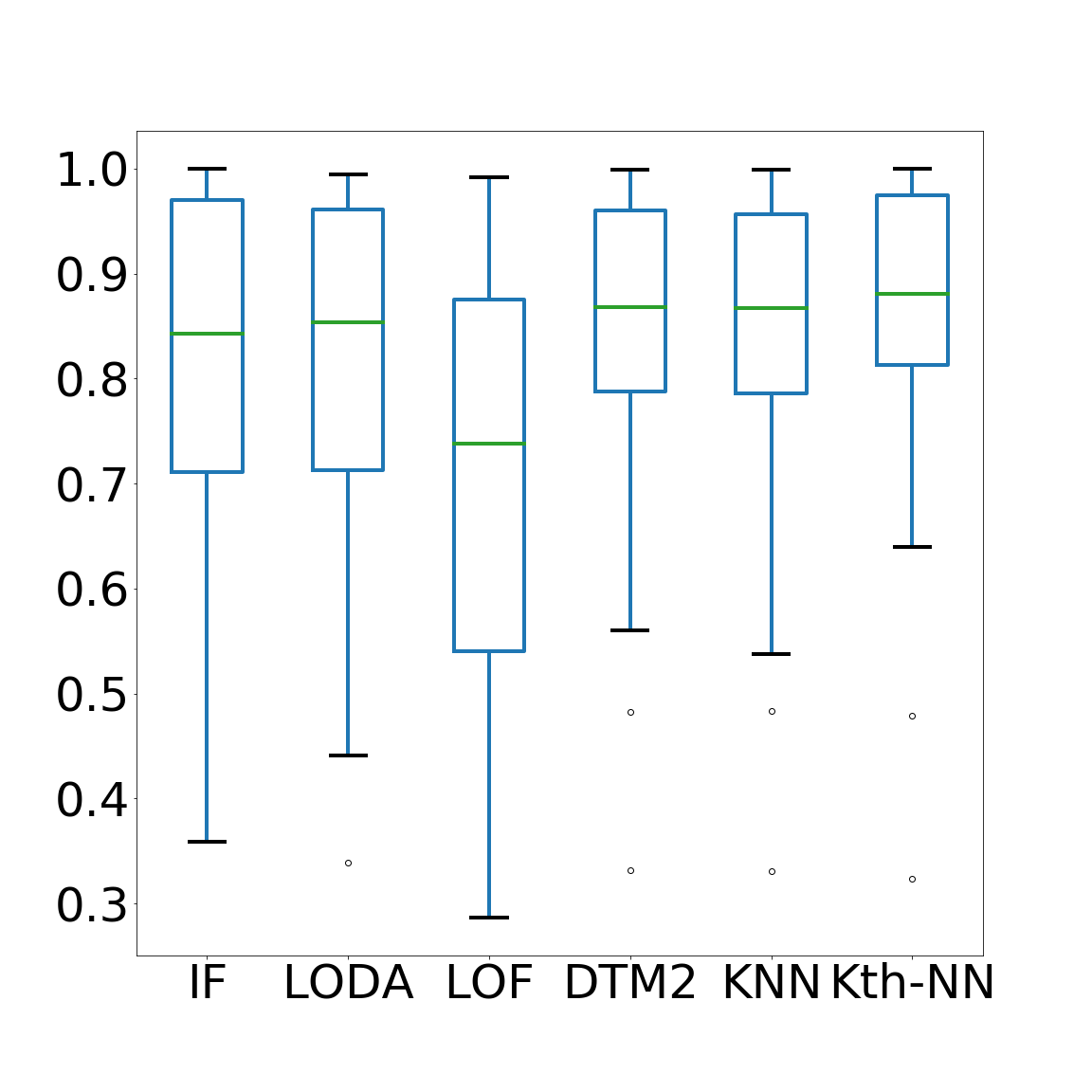}
        \caption{AUC}
    \end{subfigure}
    \begin{subfigure}[t]{0.49\textwidth}
        \centering
        \includegraphics[width=\textwidth]{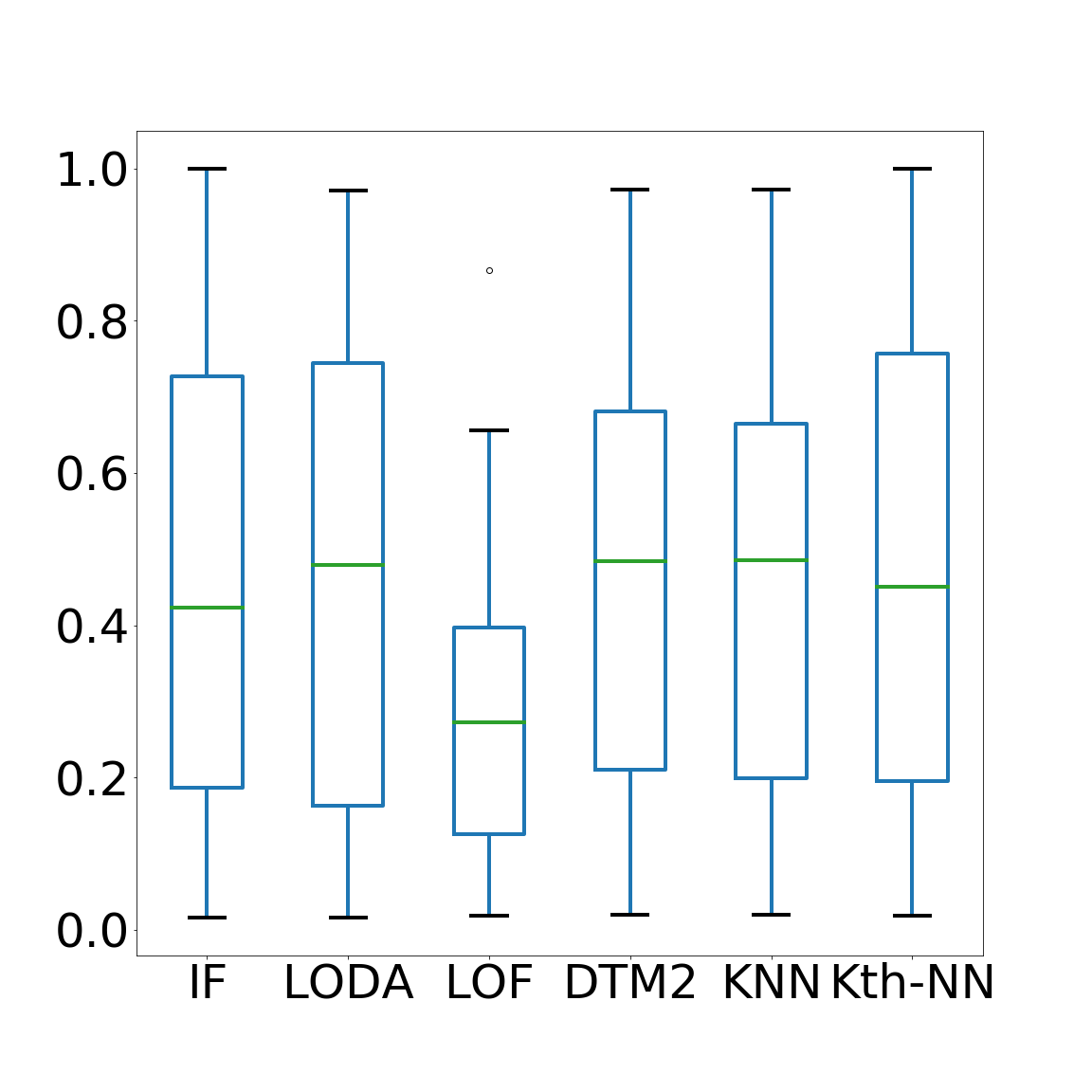}
        \caption{AP}
    \end{subfigure}
    \caption{Boxplots for AUC and AP scores on 23 real datasets.} 
    \label{box}
\end{figure}

Next, we compare the performance of $\mathrm{IForest}$, $\mathrm{LODA}$, $\mathrm{LOF}$, $\mathrm{DTM}_2$, $k\mathrm{NN}$ and $k^{\mathrm{th}}\mathrm{NN}$ on 23 real datasets from the ODDS library \cite{ODDS}. Figure \ref{box} presents the overall distributions of AUC and AP scores of the five methods as boxplots. It appears that all methods except for $\mathrm{LOF}$ have comparable performance, and we further verified this claim via pairwise Wilcoxon signed-rank tests between methods,
which showed no statistically significant difference at
level 0.05. The exact AUC and AP scores for each dataset are given in the Appendix.

\subsection{Effect of the dimension}
We then take a closer look at the performance of $\mathrm{IForest}$, $\mathrm{LODA}$, $\mathrm{LOF}$, $\mathrm{DTM}_2$, $k\mathrm{NN}$ and $k^{\mathrm{th}}\mathrm{NN}$ when the data is high dimensional. Additionally, we include the analysis of $\mathrm{DTMF}_2$ in our experiments, a quantity defined as the inverse ratio of the $\mathrm{DTM}_2$ of a point and the average $\mathrm{DTM}_2$ of its $k$-nearest neighbors. $\mathrm{DTMF}_2$ can be interpreted as a $\mathrm{LOF}$ version of $\mathrm{DTM}_2$ and is described in the Appendix. We consider six high dimensional real datasets from the UCI library \cite{UCI} (see \cite{LODA} for details) and compute the AUC and AP scores for each algorithm. The results are presented in Table \ref{tab:high}. The $n$ and $d$ columns stand for the number of samples and dimension of the datasets. On datasets \textit{gisette}, \textit{isolet} and \textit{letter}, the performance of $\mathrm{IForest}$ and $\mathrm{LODA}$ have significantly downgraded; the NN-methods give somewhat better performance, whereas $\mathrm{LOF}$ and  $\mathrm{DTMF}_2$ are showing significantly stronger performance. However, on datasets \textit{cancer} and \textit{ionoshphere}, where dimensions are slightly lower, the situations are reversed, with $\mathrm{LOF}$ and $\mathrm{DTMF}_2$ giving significantly worse performance than the others. This is consistent with our findings in Section \ref{sec:real}. The deficiency of $\mathrm{IForest}$ in high dimensions is expected, as the $\mathrm{IForest}$ trees are generated by random partitioning along a randomly selected feature. However, in high dimensions, there is a high probability that a large number of features are neglected in the process. 

\begin{table}
     \caption{AUC and AP performance on high dimensional datasets}
    \begin{subtable}[h]{1\textwidth}
        \centering
        \begin{tabular}{|c||c|c|c|c|c|c|c|c|c|}
\hline
    AUC        & $n$   & $d$   &$\mathrm{IForest}$ & $\mathrm{LODA}$ & $\mathrm{LOF}$ & $k\mathrm{NN}$ & $k^{\mathrm{th}}\mathrm{NN}$ & $\mathrm{DTM}_2$& $\mathrm{DTMF}_2$ \\ \hline \hline
gisette & 3850 &  4970  & 0.5023 & 0.5176 & 0.6753 & 0.5696 & 0.5429 & 0.5692 & 0.7051\\ 
isolet  &  4886& 617 &0.5485 & 0.5421 & 0.7330 &0.6810&0.6480 & 0.6796 & 0.7645\\ 
letter    & 4586& 617   & 0.5600 & 0.5459 & 0.7846&0.7162& 0.6826 &0.7149 &0.8096\\
madelon    &1430&  500   & 0.5327 & 0.5427 & 0.5450 &0.5608&0.5552&0.5607 &0.5546\\ 
cancer     &385&30   & 0.9528 & 0.9626 & 0.8097 &0.9780&0.9756& 0.9773&0.6937\\ 
ionosphere   &242& 33   & 0.9265 & 0.9118 & 0.9450&0.9832& 0.9803&0.9824 &0.9372\\ 
\hline
\end{tabular}
\label{high_table_auc}
       \caption{AUC}
       \label{tab:week1}
    \end{subtable}
    \hfill
    \begin{subtable}[h]{1\textwidth}
        \centering
        \begin{tabular}{|c||c|c|c|c|c|c|c|c|c|}
\hline
    AP       & $n$   & $d$   &$\mathrm{IForest}$ & $\mathrm{LODA}$ & $\mathrm{LOF}$ & $k\mathrm{NN}$ & $k^{\mathrm{th}}\mathrm{NN}$ & $\mathrm{DTM}_2$& $\mathrm{DTMF}_2$  \\ \hline \hline
gisette & 3850 &  4970  & 0.0877 & 0.0907 & 0.1628 & 0.1093 & 0.1015 & 0.1092&0.1723\\ 
isolet  &  4886& 617 &0.1005 &0.1003 & 0.2343 &0.2074&0.1846 &0.2070 &0.2458\\ 
letter    & 4586& 617   & 0.0956 & 0.0980 & 0.2921&0.2328& 0.2054 &0.2319 &0.3010\\
madelon    &1430&  500   & 0.1067 & 0.0974 & 0.1171 &0.1209&0.1181 &0.1209 &0.1166\\ 
cancer     &385&30   & 0.6274 & 0.8277 & 0.3121 &0.8813&0.8840& 0.8864&0.2800\\ 
ionosphere   &242& 33   & 0.7222 & 0.7438 & 0.6058&0.8903& 0.8801&0.8868 &0.6105\\ 
\hline
\end{tabular}
\label{high_table_ap}
        \caption{AP}
        \label{tab:week2}
     \end{subtable}
     \label{tab:high}
\end{table}

Overall, our experiments show that $\mathrm{IForest}$ and NN-methods are the top two methods with excellent overall performance on both low dimensional synthetic and real datasets. However, NN-methods exhibit better performance than $\mathrm{IForest}$ when the data is high dimensional. In the following sections, we provide a theoretical understanding of how the NN-methods work under the anomaly detection framework.


\section{Theoretical Analysis} \label{sec:theory}
In this section we formalize the settings for a simple yet natural anomaly detection problem based on the classic Huber-contamination model \cite{Huber1,huber2}, whereby a target distribution generating  normal observations is corrupted by a distribution from which anomalous observations are drawn. We introduce the notion of distance-to-a-measure (DTM) \cite{dtm_first}, as an overall functional of the data based on nearest neighbors statistics and provide finite sample bounds on the empirical nearest neighbor radii and on the rates of consistency of the DTM in the supremum norm. These theoretical guarantees are novel and may be of independent interest. Finally, we derive conditions under which DTM-based methods provably separate normal and anomalous points, as a function of the level of contamination and the separation between the normal distribution and the anomalous distribution. All the proofs are given in the
Appendix.

\subsection{Problem Setup}\label{sec:setup}
We assume  we observe $n$ i.i.d. realization $\mathbb{X}_n = (X_1,\ldots,X_n)$  from a distribution $P$ on $\mathbb{R}^d$ that follows the Huber contamination model \cite{Huber1,huber2}
\begin{align*}
 P = (1-\varepsilon)P_0 + \varepsilon P_1,
\end{align*}
where $P_0$ and $P_1$ are, respectively, the underlying distribution for the normal and anomalous instances, and $\varepsilon \in [0,1)$ is the proportion of contamination. Letting $S_0$ and $S_1$ be the support of $P_0$ and $P_1$, respectively, we further assume that $S_0 \cap S_1 = \emptyset$. The distributions $P_0$ and $P_1$, their support and  the level of contamination $\varepsilon$ are unknown. 

Our goal is devise a procedure that is able to discriminate the normal observations $X_i$'s belonging to $S_0$, from the anomalous one, falling in the set $S_1$. Since we will be focusing exclusively on NN methods, we will begin by introducing a population counterpart to the notion of $k$th nearest neighbor. Throughout the article, for any $x \in \mathbb{R}^d$ and $r>0$, $B(x,r)$ denotes the closed Euclidean ball of radius $r$ centered at $x$. 

\begin{defn}[$p$-NN radius]
Let $p \in (0,1)$. For any $x$, define $r_p(x)$ to be the radius of the smallest ball centered at $x$ with $P$-probability mass at least $p$. Formally,
$$r_{p}(x) = \inf\{r>0: P(B(x,r)) \geq p\}.$$
\end{defn} 

Setting, for a non-negative  integer $k \leq n$, $p = \frac{k}{n}$, the $k$th-nearest neighbor radius of a point $x \in \mathbb{R}^d$ with respect to the sample $(X_1,\ldots,X_n)$ is simply the $p$-NN radius $\hat{r}_p(x)$  of the corresponding empirical measure $P_n$ --  the probability measure that puts mass $1/n$ on each $X_i$. Thus,
\[
P_n(B(x,\hat{r}_p(x)) = \frac{1}{n}\left| \{ X_1,\ldots,X_n \} \cap B(x,\hat{r}_p(x)) \right| \geq \frac{k}{n}.
\]
Throughout the rest of the article, we take $p = \frac{k}{n}$.

We will impose the following, mild regularity assumptions on the distribution $P$:
\begin{itemize}
\item \textbf{Assumption (A0):}\\
$S_0$ is compact, and $S_0$ and $S_1$ are disjoint.
\item \textbf{Assumption (A1):}\\
There exists positive constants $C = C(P)$ and $\varepsilon_0 = \varepsilon_0(P)$ such that for all $0 < \varepsilon < \varepsilon_0$ and $\eta \in \mathbb{R}$,
$$|P(B(x,r_p(x)+\eta)) - P(B(x,r_p(x)))|\leq \varepsilon \Rightarrow |\eta|< C\varepsilon,$$
for $P$-almost every $x$. 
\item \textbf{Assumption (A2):}\\
$P_0$ satisfies the \textbf{(a,b)-condition}: For $b > 0$, for any $x \in S_0$, there exists $a = a(x) > 0$, and $r>0$ such that $P_0(B(x,r)) \geq \min\{1,ar^b\}$. 
\end{itemize}

Intuitively, assumption (A1) implies that $P$ has non-zero probability content around the boundary of $B(x,r_p(x))$. Observing further that the function  $r \in \mathbb{R}_+ \mapsto F_x(r) = P(B(x,r))$ is the c.d.f. of the random variable  $\| X - x\|$, where $X \sim P$, then a sufficient condition for (A1) to hold is that, uniformly over all $x$, $F_x$ has its derivative uniformly bounded away from zero in a fixed neighborhood of $r_p(x)$. This condition, originally formulated in \cite{dtm_paper} to derive bootstrap-based confidence bands for the DTM function, appears to be a natural regularity assumption in  the analysis of  NN-type methods.
When $a(x) = a$ for all $x\in S_0$, assumption  (A2) reduces to a widely used condition in the literature on statistical inference for geometric and topological data analysis \cite{JMLR:v16:chazal15a,cuevas1997}. Such condition requires  the support of $P_0$ to not locally resemble a lower dimensional set; in particular, it prevents $S_0$ from having thin ridges or outward cusps. When (A2) is violated, it becomes impossible to estimate $S_0$, no matter the size of the sample. The parameter $b$ can be interpreted as the intrinsic dimension of $P$. In particular, if $P$ admits a strictly positive density on a $D$-dimensional smooth manifold, then it can be shown that $b = D$. 

\begin{defn}[DTM \cite{dtm_first}]
The distance-to-a-measure (DTM) with respect to a probability distribution $P$ with parameter $m \in (0,1)$ and power $q \geq 1$ is defined as 
\begin{align} \label{dtm_def}
d(x) = d_{P,m,q}(x) = \left(\frac{1}{m} \int_0^m r_p(x)^q \, dp \right)^{1/q}.
\end{align}
When $q = \infty$, we set $d(x) =d_{P,m,\infty}(x) = r_m(x) $.
\end{defn}

It is immediate from the definition that a point $x \in \mathbb{R}^d$ has a small DTM value $d(x)$ if its $p$-NN radii, when averaged across all $p \in (0,m)$ are small. Intuitively, $d(x)$ can be thought of as a measure of the distance of $x$ from the bulk of the mass of the probability distribution $P$ at level of accuracy specified by the parameter $m$. The choice of the parameter $q$ allows to weight differently the impact of large versus small $p$-NN radii. 

By substituting $r_p(x)$ with $\hat{r}_p(x)$ in \eqref{dtm_def}, the empirical DTM can be seen to be  
\begin{align*}
\hat{d}(x) = d_{P_n, m,q}(x) = \left(\frac{1}{k} \sum_{X_i \in N_k(x)} \|X_i - x\|^q \right)^{1/q},
\end{align*}
where $k = \ceil{mn}$ and $N_k(x)$ denotes the set of $k$-nearest neighbors to $x$ among the sample. Different values of $q\geq 1$ yield  different NN-functionals. In particular,  the empirical DTM with $q=1$ is equivalent to the $k\mathrm{NN}$ method, and the empirical DTM with $q = \infty$ is equivalent to $k^{\mathrm{th}}\mathrm{NN}$. The notion of DTM was initially introduced in the geometric inference literature \cite{dtm_paper}, where DTM was developed for  shape reconstruction under the presence of outliers. The DTM is known to have several nice properties: it is 1-Lipschitz and it is robust with respect to perturbations of the original distributions  with respect to the Wasserstein distance. The case of $q = 2$ is special: the corresponding DTM, denoted below as $\mathrm{DTM}_2$, is also semi-concave and distance-like, and admits strong regularity conditions on its sub-level sets. Chazal et al. \cite{dtm_paper} have also derived the limiting distribution and confidence set for DTM.

\subsection{Uniform  bounds for $\hat{r}_p$ and $\hat{d}$}

In this section we derive finite sample bounds on the deviation of $\hat{r}_p$ and $\hat{d}$ from  $r_p$ and $d_{P, m,q}$, respectively, that hold uniformly over all $x \in \mathbb{R}^d$ or only over the sample points. These theoretical guarantees are, to the best of our knowledge, novel and may be of independent interest.

\begin{theorem} \label{thm_r}
Let $\delta \in (0,1)$, denote $\beta_n = \sqrt{(4/n)((d+1)\log{2n} + \log{(8/\delta)})}$. Under assumption (A1), the following bound is satisfied with probability at least $1-\delta$:
\begin{align*}
\sup_x |\hat r_p(x) - r_p(x)| \leq C(\beta_n^2 + \beta_n \sqrt{p})
\end{align*}
\end{theorem}

The dimension $d$ enters in the previous bound in such a way that, for fixed $p$, $\sup_x |\hat r_p(x) - r_p(x)| \to 0$ with high probability provided that $\frac{d}{n} \to 0$. If we limit the supremum only to the sample points, then  the dependence on the dimension disappears altogether and we can instead achieve a nearly-parametric rate of $\sqrt{ \frac{ \log n}{n}}$.

\begin{theorem}\label{cor:sample}
Let $\delta \in (0,1)$, $p = k/n$, and $p' = (k-1)/(n-1)$. \\
Denote $\alpha_n = \sqrt{(4/(n-1))(\log{2(n-1)} + \log{(8n/\delta)})}$. Under assumption A1, the following bound is satisfied with probability at least $1-\delta$:
\begin{align*}
\max_{i=1,\dots,n} |\hat r_p(X_i) - r_p(X_i)| \leq C(\alpha_n^2 + \alpha_n \sqrt{p'} + \frac{1}{n})
\end{align*}
\end{theorem}

The results in \Cref{thm_r} and \Cref{cor:sample} yield the following uniform bounds for the DTM of all order.
\begin{theorem}\label{thm:d}
Under assumption (A1), with probability at least $1-\delta$, 
\begin{align} \label{res1}
\sup_x |d(x) - \hat d(x)| \leq C\beta_n(\beta_n +\sqrt{m}),
\end{align} 
and
\begin{align} \label{res2}
\max_{i=1,\dots,n} |d(X_i) - \hat d(X_i)| \leq C\alpha_n(\alpha_n +\sqrt{m}).
\end{align} 
where $\beta_n$ and $\alpha_n$ are defined in \Cref{thm_r} and \Cref{cor:sample}.
\end{theorem}

\begin{remark}
The bound in \Cref{thm:d} holds for all choices of $q \geq 1$, including the case of $q = \infty$. Evaluating explicitly the integral $\int_0^m(\beta_n + \sqrt{p})^q \, dp$  will bring out an explicit dependence on $q$ but will not lead to better rates.
\end{remark}

\subsection{DTM for anomaly detection: theoretical guarantees}

We are now ready to derive some theoretical guarantees on the performance of DTM-based methods for discriminating normal and anomalous points in the sample $(X_1,\ldots,X_n)$ according to the Huber-contamination model described above in \Cref{sec:setup}. We recall that in our setting, a sample point $X_i$ is normal if it belongs to the support $S_0$ of $P_0$, and is otherwise deemed an anomaly if it lies in $S_1$, the support of $P_1$, where $S_1 \cap S_0 = \emptyset$. 

The methodology we consider is quite simple, and it is consistent with the prevailing practice of assigning to each sample point a score  that expresses its degree of being anomalous compared to the other points. In detail, we rank the sample points based on their empirical DTM values, and we declare the points with largest empirical DTM values as anomalies. This simple procedure will work perfectly well if 
\[
\max_{X_i \colon X_i \in S_0} \hat{d}(X_i) < \min_{x_I \in S_1}\hat{d}(X_i)
\]
and if the difference between the two quantities is large.
In general, of course, one would expect that some sample points in $S_0$ may have smaller empirical DTMs of some of the points in $S_1$. The extent to which such incorrect labeling occurs depends on two key factors: how closely the empirical DTM tracks the true DTM and whether the population DTM could itself discriminate normal points versus anomalous ones. The former issue can be handled using the high probability bounds on the stochastic fluctuations of the empirical DTM obtained in the previous section. The latter issue will instead require to specify some degree of separation between the mixture components $P_0$ and $P_1$, both in terms the distance between their supports but also in terms of how their probability mass gets distributed. There is more than one way to formalize this setting. Here we choose to remain completely agnostic to the form of the contaminating distribution $P_1$, for which we impose virtually no constraint. On the other hand, we require the normal distribution $P_0$ to satisfy condition (A2) above in such a way that point inside the support will have larger values of $a(x)$ than points near the boundary of $S_0$. This condition, which is satisfied if for example $P_0$ admits a Lebesgue density whose values increase as a function of the distance from the boundary of $S_0$, ensures that the population DTM will be large near the boundary of $S_0$ and small everywhere else. As a result, incorrect labeling of normal points will only occur around the boundary of $S_0$ but not inside the bulk the distribution $P_0$. We formalize this intuition in our next result, which is purely deterministic.

\begin{prop} \label{prop}
Under assumptions (A0) and (A2), suppose that $a(x) = g(d(x,\partial S_0))$, where $g(z)$ is a non-decreasing function on $[0,z_0)$ for some $z_0$, and $g(z) \geq g(z_0)$ for all $z \geq z_0$. Let 
\begin{equation}\label{eq:eta}
\eta = \min_{x \in S_0, y \in S_1} \| x - y\|
\end{equation}
be the distance between $S_0$ and $S_1$ and $h>0$ be a given threshold parameter.  For any $m > \varepsilon$, additionally assume that 
\begin{equation}\label{eq:g0}
g(z_0) \geq g_0 :=
\begin{cases}
\frac{m}{1-\varepsilon}\left(\frac{b+q}{b}\left(\frac{m-\varepsilon}{m} \eta^q - h\right)\right)^{-b/q} & 1 \leq  q < \infty \\
\frac{m}{1-\varepsilon}(\eta - h)^{-b} & q = \infty.
\end{cases}
\end{equation}
Next, define the "safety zone" $A_\eta$ as
\begin{equation}\label{eq:Aeta}
A_\eta = \left\{x \in S_0: d(x,\partial S_0) \geq g^{-1}(g_0)\right\} 
\end{equation}
Then, we have
\begin{align} \label{eq:main}
\sup_{x \in A_\eta} d_{P,m,q}(x) + h < \inf_{y \in S_1} d_{P,m,q}(y).
\end{align}
\end{prop}

The main message from the previous result is that there exists a subset $A_\eta$ of the support of the normal distribution, which intuitively corresponds to a region deep inside the support of $P_0$ of high density, over which the population DTM will be smaller than at any point in the support $S_1$ of the contaminating distribution. Thus, the true DTM is guaranteed to perfectly separate $A_\eta$ from $S_1$, making mistakes (possibly) only for the normal points in $S_0 \setminus A_{\eta}$. 

Notice that the definition of $A_\eta$ depends on all the relevant quantities, namely the contamination parameter $\varepsilon$, the probability parameter $m$, the dimension $b$ of $P_0$ and the order $q$ of the DTM through the expression \eqref{eq:g0}.
Importantly, it is necessary that $m > \varepsilon$, otherwise inequality \eqref{eq:main} maybe not be satisfied. For example, we can take $P_1$ to have point mass at a single point $y$; then $r_{P,t}(y) = 0$ for all $t \leq m$, and the right hand side of \eqref{eq:main} is zero. 

When $g(0) = a_0 > 0$, which occurs, e.g., if $P_0$ has a density bounded away from $0$ over its support, implies that $A_\eta = S_0$ if 
\begin{align*}
\eta > \left(\frac{m}{m-\varepsilon}\left(\frac{b}{b+q} \left(\frac{m}{a_0(1-\varepsilon)}\right)^{q/b} + h\right) \right)^{-1/q}.
\end{align*}
That is, when $S_0$ and $S_1$ are sufficiently well-separated, the DTM will classify all the points in $S_0$ as normals.

The parameter $h$ serves as a buffer that allows one to replace the DTM function $d(x)$ with any estimator that is close to it  in the supremum norm by no more than $h$. Thus, we may plug-in the high-probability bounds of \Cref{cor:sample} and \Cref{thm_r} to conclude that the empirical DTM will will identify all normal instances within $A_\eta$ correctly, with high probability.

\begin{corollary}
Taking $h$ to be twice the upper bound in \eqref{res2}, we get, with probability at least $ 1- \delta$,
\begin{align*}
\max_{X_i \in A_\eta} \hat d_{P,m,q}(X_i) < \min_{X_i \in S_1} \hat d_{P,m,q}(X_i).
\end{align*}
Similarly, if $h$ is twice the upper bound in \eqref{res1}, we have that
\begin{equation}\label{eq:other.result}
\sup_{x \in A_\eta} \hat d_{P,m,q}(x) < \inf_{y \in S_1} \hat d_{P,m,q}(y).
\end{equation}
\end{corollary}

The guarantee in \eqref{eq:other.result} calls for a higher sample complexity that depends on the dimension $d$. At the same time, it extends to all the points in $A_\eta$ and not just the sample points. Thus the DTM can accurately identify not only the normal instance in the sample but any other normal instance, such as future observations.

\subsection{Illustrative examples}
\begin{figure}[H]
    \centering
    \begin{subfigure}[t]{0.32\textwidth}
        \centering
        \includegraphics[width=\textwidth]{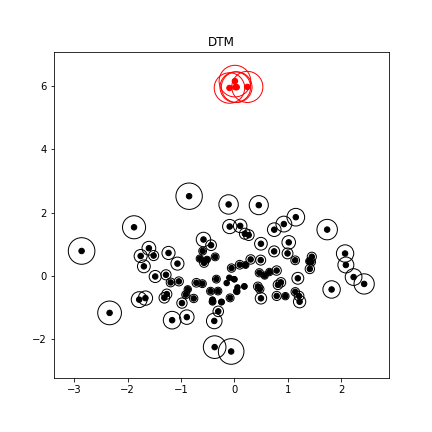}
        \caption{High Separation}
    \end{subfigure}
    \begin{subfigure}[t]{0.32\textwidth}
        \centering
        \includegraphics[width=\textwidth]{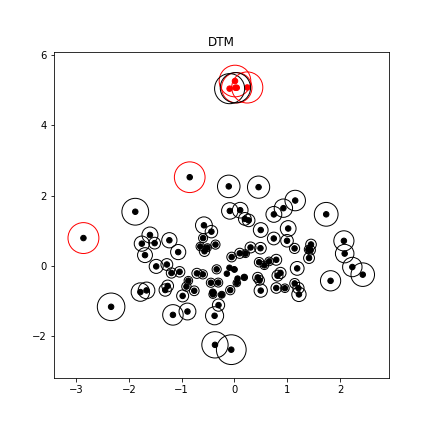}
        \caption{Medium Separation}
    \end{subfigure}
    \begin{subfigure}[t]{0.32\textwidth}
        \centering
        \includegraphics[width=\textwidth]{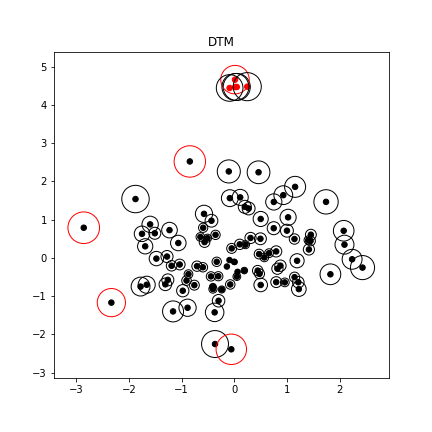}
        \caption{Low Separation}
    \end{subfigure}
    \caption{Performance of DTM when the separation distance between the normal instances and anomalies gradually decreases.} 
    \label{demo}
\end{figure}

We illustrate the separation condition in Proposition \ref{prop} with the following example. Consider a collection of normal instances generated from a standard normal distribution. Figure \ref{demo} shows the mis-classification rates for $\mathrm{DTM_2}$ as a cluster of 5 anomalies approaches the normal instances. The color of each point represents its class, with black being the normal instances and red being anomalies. The radius of the circle around each point represents its empirical DTM score, and the color of the circle represents its predicted class from $\mathrm{DTM_2}$. As we see, as the anomalies approach the normal instances, more and more data around the boundaries of the normal distribution get mis-classified as anomalies.

\section{Conclusions}
In this paper we have presented  empirical evidence, based on simulated and real-life benchmark datasets, that  NN-based methods show very good performance at identifying anomalous instances in an unsupervised anomaly detection set-up. We have introduced a simple but natural framework for anomaly detection based on the Huber contamination model and have used it to characterize the performance of a class of NN methods for anomaly detection that are based on the distance-to-a-measure (DTM) functional. In our results we rely on various geometric and analytic properties of the underlying distribution to the  accuracy of DTM-methods for anomaly detection. We are able to demonstrate that, under mild conditions, NN methods will mis-classify normal points only around the boundary of the support of the distribution generating normal instances and have quantified this phenomenon rigorously. Finally, we have derived novel finite sample bounds on the nearest neighbor radii and on the rate of convergence of the empirical DTM to the true DTM that may be of independent interest.


\appendix
\appendixpage

\section{Definition for $\mathrm{DTMF}_2$}
\begin{defn}
The $\mathrm{DTM}_2$ and $\mathrm{DTMF}_2$ scores are defined as:
\begin{align*}
\mathrm{DTM}_2(x) = \left(\frac{1}{k} \sum_{X_i \in N_k(x)} \|X_i - x\|^2 \right)^{1/2},
\end{align*}
\begin{align*}
\mathrm{DTMF}_2(x) = \frac{1}{|N_k(x)|}\sum_{y \in N_k(x)}\frac{\mathrm{DTM}_2(y)}{\mathrm{DTM}_2(x)}.
\end{align*}
\end{defn}

\section{Proof of Theorems}
\subsection{Proof of Theorem 3.3}
\begin{proof}
By standard VC theory \cite{vc1,vc2}, for any ball $B \subset \mathbb{R}^d$, we have
\begin{align}\label{eq1}
P(B) \geq p + \beta_n^2 + \beta_n \sqrt{p} \Rightarrow P_n(B) \geq p.
\end{align}
\begin{align}\label{eq2}
P(B) < p - \beta_n^2 - \beta_n \sqrt{p} \Rightarrow P_n(B) < p
\end{align}
with probability at least $1-\delta$.

\textit{Step1:} First, we want to show that
\begin{align}\label{s1}
\hat r_p(x) \leq r_p(x) +  C(\beta_n^2 + \beta_n \sqrt{p})
\end{align}
for all $x$. By definition of $r_p(x)$, we have $P(B(x,r_p(x))) \geq p$. Define $r^+ = \inf\{r: P(B(x,r)) \geq p + \beta_n^2 + \beta_n \sqrt{p}\}$. Then, we have
\begin{align*}
P(B(x,r^+)) \geq p + \beta_n^2 + \beta_n \sqrt{p} \Rightarrow P_n(B(x,r^+)) \geq p
\end{align*}
by \eqref{eq1}. Therefore, $r^+ \geq \hat r_p(x)$. Next, note that $r_p(x) \leq r^+$. If $r_p(x) = r^+$, \eqref{s1} holds trivially. If $r_p(x) < r^+$, then for all $s$ such that $r_p(x) < s < r^+$, we have
\begin{align*}
p \leq P(B(x,r_p(x))) \leq P(B(x,s)) \leq p + \beta_n^2 + \beta_n \sqrt{p}.
\end{align*}
Then by assumption (A1), 
\begin{align*}
s \leq r_p(x) + C(\beta_n^2 + \beta_n \sqrt{p}).
\end{align*}
Taking $s \uparrow r^+$, we get $\hat r_p(x) \leq r^+ \leq r_p(x) + C(\beta_n^2 + \beta_n \sqrt{p})$ as desired.\\

\textit{Step 2:} Next, we want to show the reverse direction:
\begin{align}\label{s2}
r_p(x) \leq \hat r_p(x) + C(\beta_n^2 + \beta_n \sqrt{p}).
\end{align}
Let $r^- = \inf \{r: P(B(x,r)) \geq p - \beta_n^2 - \beta_n \sqrt{p}\}$. Then, clearly $r^- \leq r_p(x)$ and $P(B(x,r^-)) \geq p - \beta_n^2 - \beta_n \sqrt{p}$. For all $s < r^-$, we have
\begin{align*}
&P(B(x,s)) < p - \beta_n^2 - \beta_n\sqrt{p}\\
&\Rightarrow  P_n(B(x,s)) < p\\
&\Rightarrow s < \hat r_p(x)
\end{align*}
where the first implication follows from \eqref{eq2}. Taking $s \uparrow r^-$, we get $r^- \leq \hat r_p(x)$. If $r^- =  r_p(x)$, \eqref{s2} holds trivially. If $r^- < r_p(x)$, then for any $u$ satisfying $r^- < u < r_p(x)$, we have
\begin{align*}
&p - \beta_n^2 - \beta_n \sqrt{p} \leq P(B(x,r^-)) \leq P(B(x,u)) \leq p\\
&\Rightarrow u \leq r^- + C(\beta_n^2 + \beta_n \sqrt{p}).
\end{align*}
Taking $u \uparrow r_p(x)$, we get
\begin{align*}
r_p(x) \leq r^- + C(\beta_n^2 + \beta_n \sqrt{p}) \leq \hat r_p(x) + C(\beta_n^2 + \beta_n \sqrt{p})
\end{align*}
as desired.
\end{proof}

\subsection{Proof of Theorem 3.4}
\begin{proof}
By assumption (A1), it suffices to show that
\begin{align*}
    \max_{i=1,\dots,n} |P(B(x_i,r_p(x_i))) - P(B(x_i,\hat r_p(x_i)))| \leq \alpha_n^2 + \alpha_n \sqrt{p'} + \frac{1}{n}
\end{align*}
with probability higher than $1-\delta$.

Fix $X_i$, and let $P_{i,n-1}$ be the marginal distribution of $\mathbb{X}_n \backslash \{X_i\}$. Define the event 
$$A_{i,n} = \{|P(B(X_i,\hat r_p(X_i))) - P_{i,n-1}(B(X_i,\hat r_p(X_i)))| > \alpha_n^2 + \alpha_n \sqrt{p'}\},$$
and thus 
\begin{align*}
    \max_{i=1,\dots,n} |P(B(X_i,\hat r_p(X_i))) - P_{i,n-1}(B(X_i,\hat r_p(X_i)))| > \alpha_n^2 + \alpha_n \sqrt{p'} = \bigcup_{i=1}^n A_{i,n}.
\end{align*}
Then, $A_{i,n}$ is contained in the event 
$$B_{i,n} = \{|P(B(X_i,r)) - P_{i,n-1}(B(X_i, r))| > \alpha_n^2 + \alpha_n \sqrt{p'}, \, \forall r>0\}.$$
We have, 
\begin{align*}
    P(\bigcup_{i=1}^n A_{i,n}) \leq P(\bigcup_{i=1}^n B_{i,n}) \leq \sum_{i=1}^n \mathbb{E}_{i}[ \mathbb{E}_{-i}[1_{B_{i,n}(X_1,\dots,X_n)}]]
\end{align*}
where $1_{B_{i,n}}(\cdot)$ is the indicator function of $B_{i,n}$, $\mathbb{E}_i$ and $\mathbb{E}_{-i}$ are respectively the expectation with respect to the marginal distribution of $X_i$ and $\mathbb{X}_n \backslash \{X_i\}$.

Since for each fixed realization of $X_i$, the class of balls $B(x_i,\cdot)$ centered at $x_i$ with arbitrary radius has VC dimension 1, by standard VC theory \cite{vc1,vc2}, we have
\begin{align*}
    \mathbb{E}_{-i}[1_{B_{i,n}(X_1,\dots,X_{i-1},x_i,X_{i+1},\dots,X_n)}] \leq \frac{\delta}{n}.
\end{align*}
Therefore,
\begin{align*}
    \max_{i=1,\dots,n} |P(B(X_i,\hat r_p(X_i))) - P_{i,n-1}(B(X_i,\hat r_p(X_i)))| \leq \alpha_n^2 + \alpha_n \sqrt{p'} 
\end{align*}
with probability at least $1-\delta$. Next, since
\begin{align*}
    |P_{i,n-1}(B(X_i,\hat r_p(X_i))) - P_n(B(X_i,\hat r_p(X_i)))|  = |p'-p| \leq \frac{1}{n},
\end{align*}
\begin{align*}
    \max_{i=1,\dots,n} |P(B(X_i,\hat r_p(X_i))) - P_n(B(X_i,\hat r_p(X_i)))| \leq \alpha_n^2 + \alpha_n \sqrt{p'} + \frac{1}{n}
\end{align*}
with probability at least $1-\delta$. Finally, notice that assumption (A1) ensures that the c.d.f of the random variable $\|X - x\|$ is almost surely continuous at $r_p(x)$. Thus, we have $P_n(B(X_i,\hat r_p(X_i))) = p = P(B(X_i,r_p(X_i)))$.
\end{proof}

\subsection{Proof of Theorem 3.5}
\begin{proof}
With probability at least $1-\delta$, 
\begin{align*}
\sup_x |d(x) - \hat d(x)| &= \sup_x \left| \left(\frac{1}{m} \int_0^m r_p(x)^q \, dp\right)^{1/q} - \left(\frac{1}{m} \int_0^m \hat r_p(x)^q \, dp\right)^{1/q}\right|\\
&\leq \sup_x \left(\frac{1}{m} \int_0^m (r_p(x) - \hat r_p(x))^q \, dp \right)^{1/q} & {\text{By Minkowski Inequality}}\\
&\leq C \beta_n (\frac{1}{m} \int_0^m(\beta_n + \sqrt{p})^q \, dp)^{1/q} & {\text{By Theorem 3.3}}\\
&\leq C\beta_n (\beta_n + \sqrt{m}) \left(\frac{1}{m} \int_0^m  dp \right)^{1/q}\\
&= C\beta_n(\beta_n +\sqrt{m}).
\end{align*}
\end{proof}

\subsection{Proof of Proposition 3.6}
\begin{proof}
Equivalently, by the definition of DTM, we will need to show that
\begin{align} \label{eq}
\sup_{x \in A_\eta} \frac{1}{m} \int_0^m r_{P,t}(x)^q \, dt < \inf_{y \in S_1} \frac{1}{m} \int_0^m r_{P,t}(y)^q \, dt - h.
\end{align}
Since for any $x \in A_\eta$, $P(B(x,r)) \geq (1-\varepsilon) P_0(B(x,r))$, we have $r_{P,t}(x) \leq r_{(1-\varepsilon)P_0,t}(x) = r_{P_0,t/1-\varepsilon}(x)$. When $r = (\frac{t}{a(x)(1-\varepsilon)})^{1/b}$, by the $(a,b)$-condition of $P_0$, we have $P_0(B(x,r)) \geq a(x)r^b = \frac{t}{1-\varepsilon}$. Hence, $r_{P_0,t/1-\varepsilon}(x) \leq (\frac{t}{a(x)(1-\varepsilon)})^{1/b}$. Putting the inequalities together, the LHS of \eqref{eq} gives
\begin{align*}
\frac{1}{m} \int_0^m r_{P,t}(x)^q \, dt &\leq \frac{1}{m} \int_0^m (\frac{t}{a(x)(1-\varepsilon)})^{q/b}\, dt \\
&= \frac{b}{b+q} (\frac{m}{a(x)(1-\varepsilon)})^{q/b}
\end{align*}
Next, consider the right hand side of \eqref{eq}. We have that
\begin{align*}
\inf_{y \in S_1} \frac{1}{m} \int_0^m r_{P,t}(y)^q \, dt&= \inf_{y \in S_1} \frac{1}{m} \int_0^\varepsilon r_{P,t}(y)^q \, dt + \frac{1}{m}\int_\varepsilon^m r_{P,t}(y)^q \, dt\\
&\geq 0 + \frac{1}{m}\int_\varepsilon^m \eta^q \, dt\\
&= \frac{m-\varepsilon}{m}\eta^q
\end{align*}
Hence, \eqref{eq} holds if 
\begin{align} \label{eqq}
&\frac{b}{b+q} \left(\frac{m}{a(x)(1-\varepsilon)}\right)^{q/b} < \frac{m-\varepsilon}{m}\eta^q - h \\
&\Leftrightarrow a(x) > \frac{m}{1-\varepsilon}\left(\frac{b+q}{b}\left(\frac{m-\varepsilon}{m} \eta^q - h\right)\right)^{-b/q}\\
&\Leftrightarrow x \in A_\eta
\end{align}
\end{proof}

\section{Simulation Results on 23 Real Datasets from ODDS}
Table \ref{big_table} gives the exact AUC and AP scores of $\mathrm{IForest}$, $\mathrm{LODA}$, $\mathrm{LOF}$, $\mathrm{DTM}_2$, $k\mathrm{NN}$, and $k^{\mathrm{th}}\mathrm{NN}$ on 23 real datasets from the ODDS library. 

\begin{table}
\caption{Performance of $\mathrm{IForest}$, $\mathrm{LODA}$, $\mathrm{LOF}$, $\mathrm{DTM}_2$, $k\mathrm{NN}$, and $k^{\mathrm{th}}\mathrm{NN}$ on 23 real datasets from the ODDS library.}
\label{big_table}
\begin{subtable}[h]{1\textwidth}
\begin{tabular}{|c||c|c|c|c|c|c|}
\toprule AUC &        $\mathrm{IForest}$ & $\mathrm{LODA}$ &    $\mathrm{LOF}$  &   $\mathrm{DTM}_2$ &   $k\mathrm{NN}$ &  $k^{\mathrm{th}}\mathrm{NN}$\\
\midrule annthyroid  &  0.846217 &  0.711716 &  0.688763 &  0.677126 &  0.681196 &  0.662250 \\
arrhythmia  &  0.774180 &  0.789645 &  0.763778 &  0.807466 &  0.806681 &  0.815473 \\
breastw     &  0.988089 &  0.987891 &  0.376371 &  0.980041 &  0.979805 &  0.982081 \\
cardio      &  0.925666 &  0.904219 &  0.705637 &  0.831097 &  0.820695 &  0.880306 \\
glass       &  0.706775 &  0.771816 &  0.737669 &  0.867751 &  0.867209 &  0.869106 \\
ionosphere  &  0.842363 &  0.853369 &  0.899506 &  0.928007 &  0.928148 &  0.920141 \\
letter      &  0.600280 &  0.622487 &  0.842000 &  0.856193 &  0.861893 &  0.809837 \\
lympho      &  1.000000 &  0.992958 &  0.981221 &  0.977700 &  0.977700 &  0.978286 \\
mammography &  0.853864 &  0.866368 &  0.819344 &  0.850100 &  0.850604 &  0.849169 \\
mnist       &  0.792829 &  0.595506 &  0.839678 &  0.862295 &  0.861369 &  0.861813 \\
musk        &  0.999944 &  0.994193 &  0.286222 &  0.957031 &  0.936976 &  1.000000 \\
optdigits   &  0.714978 &  0.714282 &  0.612373 &  0.560559 &  0.537313 &  0.842404 \\
pendigits   &  0.961689 &  0.950902 &  0.850733 &  0.958278 &  0.950210 &  0.970528 \\
pima        &  0.675037 &  0.618657 &  0.557993 &  0.636045 &  0.634418 &  0.639545 \\
satellite   &  0.686132 &  0.725766 &  0.578879 &  0.768331 &  0.764688 &  0.795738 \\
satimage-2  &  0.993326 &  0.994631 &  0.991675 &  0.999054 &  0.999079 &  0.998954 \\
shuttle     &  0.997529 &  0.992264 &  0.522135 &  0.989215 &  0.984996 &  0.993954 \\
speech      &  0.441678 &  0.441248 &  0.478689 &  0.482781 &  0.483310 &  0.478594 \\
thyroid     &  0.978939 &  0.954587 &  0.963042 &  0.946970 &  0.947420 &  0.943083 \\
vertebral   &  0.359048 &  0.338889 &  0.495714 &  0.331746 &  0.330794 &  0.323968 \\
vowels      &  0.739488 &  0.757411 &  0.937155 &  0.961067 &  0.963144 &  0.946216 \\
wbc         &  0.943177 &  0.958517 &  0.910764 &  0.948113 &  0.946379 &  0.949980 \\
wine        &  0.776471 &  0.963025 &  0.428151 &  0.994958 &  0.993277 &  0.996218 \\
\bottomrule
\end{tabular}
\caption{AUC}
\end{subtable}
\begin{subtable}[h]{1\textwidth}
\begin{tabular}{|c||c|c|c|c|c|c|}
\toprule AUC &        $\mathrm{IForest}$ & $\mathrm{LODA}$ &    $\mathrm{LOF}$  &   $\mathrm{DTM}_2$ &   $k\mathrm{NN}$ &  $k^{\mathrm{th}}\mathrm{NN}$\\
\midrule annthyroid  &  0.336719 &  0.221278 &  0.252282 &  0.201405 &  0.203313 &  0.191132 \\
arrhythmia  &  0.422741 &  0.479021 &  0.372709 &  0.491718 &  0.489785 &  0.511596 \\
breastw     &  0.972689 &  0.970735 &  0.272824 &  0.945230 &  0.944475 &  0.951773 \\
cardio      &  0.577570 &  0.579294 &  0.202455 &  0.404516 &  0.399020 &  0.450174 \\
glass       &  0.096007 &  0.140315 &  0.193538 &  0.162208 &  0.162824 &  0.155266 \\
ionosphere  &  0.794018 &  0.794706 &  0.866694 &  0.928604 &  0.928993 &  0.911973 \\
letter      &  0.089123 &  0.090754 &  0.334208 &  0.260399 &  0.268795 &  0.200453 \\
lympho      &  1.000000 &  0.835714 &  0.655556 &  0.723611 &  0.723611 &  0.695202 \\
nmammography &  0.193517 &  0.264330 &  0.130258 &  0.167475 &  0.169236 &  0.161568 \\
mnist       &  0.267129 &  0.143844 &  0.397000 &  0.404172 &  0.403502 &  0.387391 \\
musk        &  0.998328 &  0.881940 &  0.021850 &  0.618577 &  0.496054 &  1.000000 \\
noptdigits   &  0.051332 &  0.047997 &  0.033582 &  0.032489 &  0.031128 &  0.081907 \\
npendigits   &  0.328479 &  0.263207 &  0.077311 &  0.217698 &  0.193527 &  0.315036 \\
npima        &  0.506879 &  0.491468 &  0.391361 &  0.486558 &  0.485157 &  0.492184 \\
satellite   &  0.659824 &  0.693257 &  0.406871 &  0.639164 &  0.634576 &  0.680913 \\
satimage-2  &  0.936035 &  0.911899 &  0.516222 &  0.972246 &  0.972113 &  0.972834 \\
shuttle     &  0.983694 &  0.825359 &  0.261309 &  0.746971 &  0.694083 &  0.818767 \\
speech      &  0.016421 &  0.015298 &  0.018916 &  0.019062 &  0.019088 &  0.018781 \\
thyroid     &  0.595528 &  0.276667 &  0.397719 &  0.297644 &  0.296979 &  0.285007 \\
vertebral   &  0.094209 &  0.090683 &  0.121829 &  0.089739 &  0.089664 &  0.088901 \\
vowels      &  0.179951 &  0.181478 &  0.396334 &  0.484752 &  0.501906 &  0.403366 \\
wbc         &  0.588631 &  0.640832 &  0.279934 &  0.495254 &  0.488687 &  0.554438 \\
wine        &  0.192461 &  0.544417 &  0.072027 &  0.941540 &  0.928312 &  0.954040 \\
\bottomrule
\end{tabular}

\caption{AP}
\end{subtable}
\end{table}

\section{Performance on the Difficult Examples}

\begin{figure}
    \centering
    \begin{subfigure}[t]{0.325\textwidth}
        \centering
        \includegraphics[width=\textwidth]{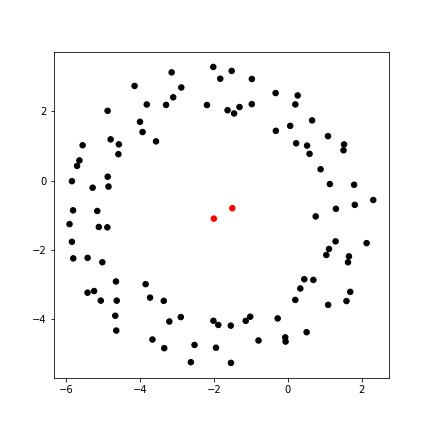}
        \caption{Ring}
        \label{fig:ring}
    \end{subfigure}
    \begin{subfigure}[t]{0.325\textwidth}
        \centering
        \includegraphics[width=\textwidth]{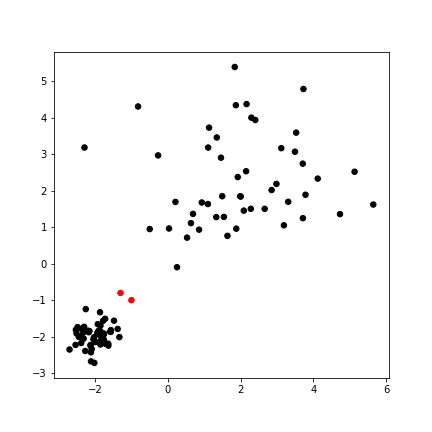}
        \caption{Local Anomalies}
        \label{fig:local}
    \end{subfigure}
    \begin{subfigure}[t]{0.325\textwidth}
        \centering
        \includegraphics[width=\textwidth]{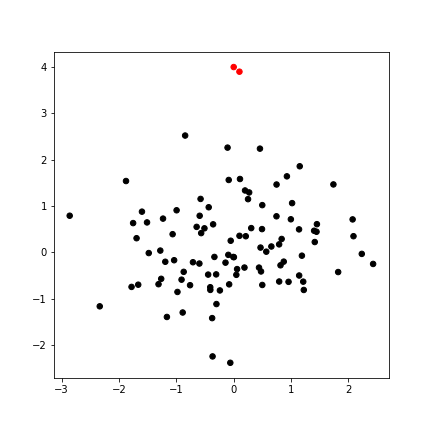}
        \caption{Clustered Anomalies}
        \label{fig:cluster}
    \end{subfigure}
    \caption{Examples of difficult datasets.}
    \label{fig:ex}
\end{figure}

Figure \ref{fig:ex} gives three examples of difficult situations where some algorithms will very likely fail. The black dots represent the normal instances, and the two red dots represent anomalies. In Figure \ref{fig:ring} where the anomalies are located in the center of a circle of normal points, $\mathrm{IForest}$ and $\mathrm{LODA}$ will have a hard time detecting the anomalies, whereas $\mathrm{LOF}$ and NN-methods have no trouble. In Figure \ref{fig:local}, if the anomalies are locally relatively far away from a group of normal points, NN-methods, $\mathrm{IForest}$, and $\mathrm{LODA}$ won't be able to pick them up, whereas $\mathrm{LOF}$ is designed to handle this specific case. However, we observed through extensive simulations that $\mathrm{LOF}$ can easily make mistakes on global anomalies, and Figure \ref{fig:cluster} gives one such example. If we have a cluster of anomalies located at some distance from a collection of normal points, $\mathrm{LOF}$ tends to mis-identify some of the anomalies as normal points, whereas the other methods have no such problem. 

\begin{figure}
    \centering
    \begin{subfigure}[t]{0.49\textwidth}
        \centering
        \includegraphics[width=\textwidth]{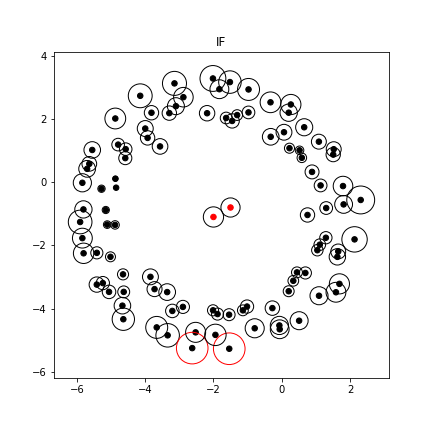}
        \caption{$\mathrm{IForest}$}
    \end{subfigure}
    \begin{subfigure}[t]{0.49\textwidth}
        \centering
        \includegraphics[width=\textwidth]{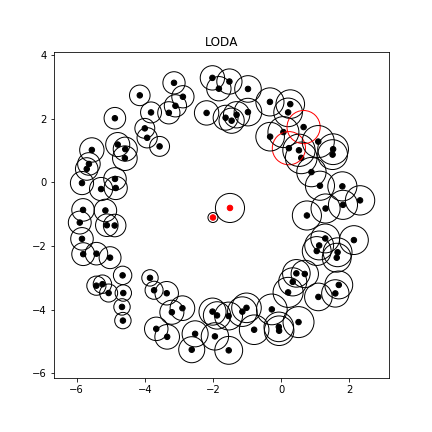}
        \caption{$\mathrm{LODA}$}
    \end{subfigure}
    \begin{subfigure}[t]{0.49\textwidth}
        \centering
        \includegraphics[width=\textwidth]{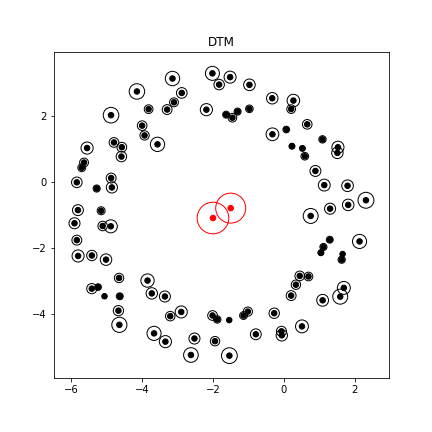}
        \caption{$\mathrm{DTM}_2$}
    \end{subfigure}
    \begin{subfigure}[t]{0.49\textwidth}
        \centering
        \includegraphics[width=\textwidth]{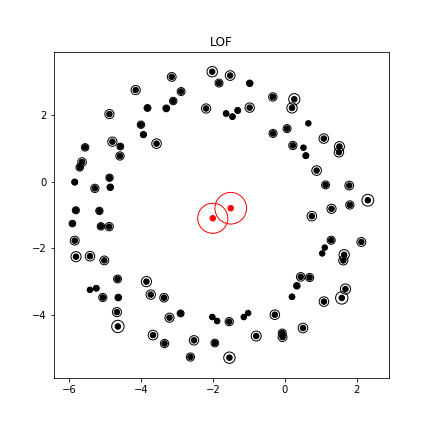}
        \caption{$\mathrm{LOF}$}
    \end{subfigure}
    \caption{Performance on the difficult datasets. Case: ring} 
    \label{fig:1}
\end{figure}

\begin{figure}
    \centering
    \begin{subfigure}[t]{0.49\textwidth}
        \centering
        \includegraphics[width=\textwidth]{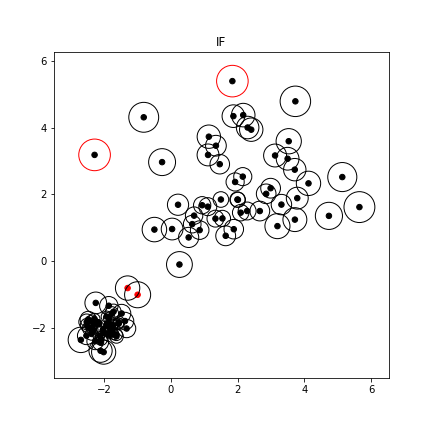}
        \caption{$\mathrm{IForest}$}
    \end{subfigure}
    \begin{subfigure}[t]{0.49\textwidth}
        \centering
        \includegraphics[width=\textwidth]{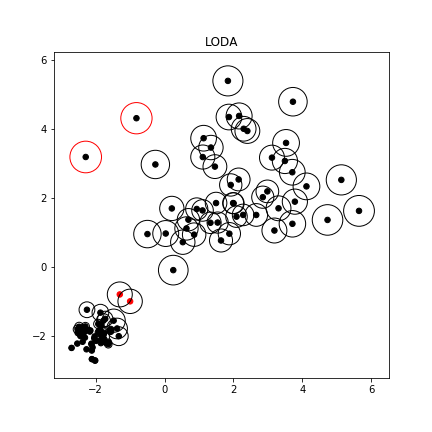}
        \caption{$\mathrm{LODA}$}
    \end{subfigure}
    \begin{subfigure}[t]{0.49\textwidth}
        \centering
        \includegraphics[width=\textwidth]{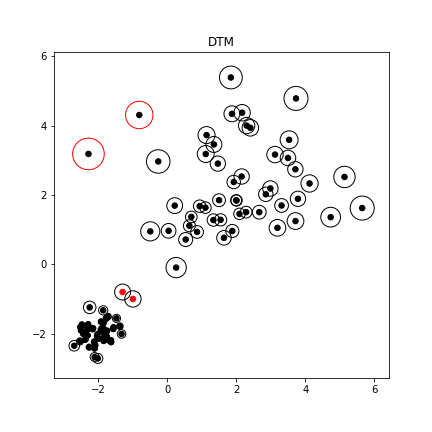}
        \caption{$\mathrm{DTM}_2$}
    \end{subfigure}
    \begin{subfigure}[t]{0.49\textwidth}
        \centering
        \includegraphics[width=\textwidth]{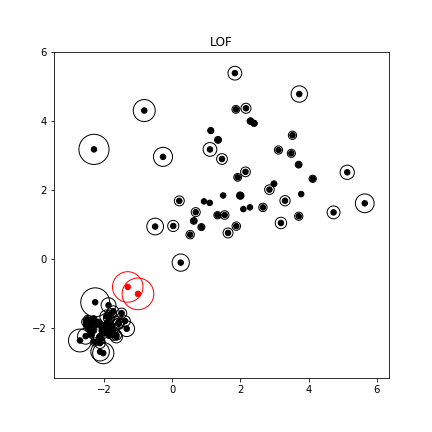}
        \caption{$\mathrm{LOF}$}
    \end{subfigure}
    \caption{Performance on the difficult datasets. Case: local anomalies} 
    \label{fig:2}
\end{figure}

\begin{figure}
    \centering
    \begin{subfigure}[t]{0.49\textwidth}
        \centering
        \includegraphics[width=\textwidth]{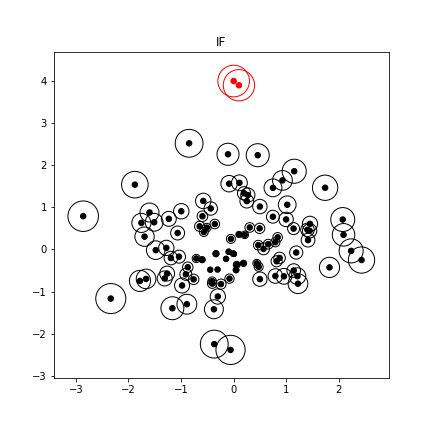}
        \caption{$\mathrm{IForest}$}
    \end{subfigure}
    \begin{subfigure}[t]{0.49\textwidth}
        \centering
        \includegraphics[width=\textwidth]{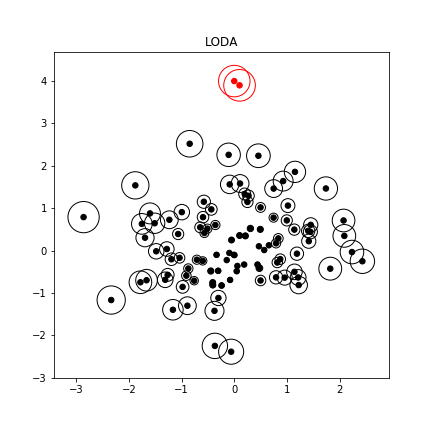}
        \caption{$\mathrm{LODA}$}
    \end{subfigure}
    \begin{subfigure}[t]{0.49\textwidth}
        \centering
        \includegraphics[width=\textwidth]{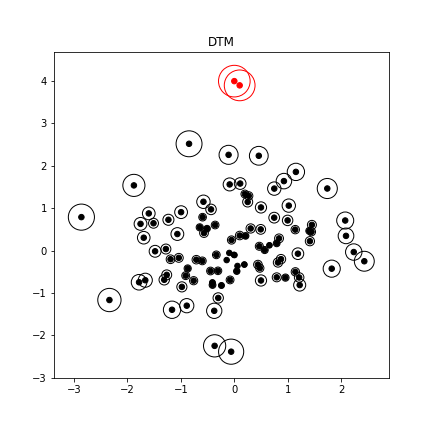}
        \caption{$\mathrm{DTM}_2$}
    \end{subfigure}
    \begin{subfigure}[t]{0.49\textwidth}
        \centering
        \includegraphics[width=\textwidth]{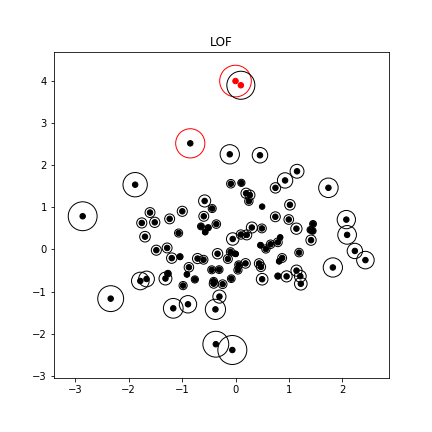}
        \caption{$\mathrm{LOF}$}
    \end{subfigure}
    \caption{Performance on the difficult datasets. Case: clustered anomalies} 
    \label{fig:3}
\end{figure}

Figure \ref{fig:1} \ref{fig:2} \ref{fig:3} show the performance of $\mathrm{IForest}$, $\mathrm{LODA}$, $\mathrm{LOF}$, and $\mathrm{DTM}_2$ in each of the difficult examples. The radius of the circle around each point gives the anomaly score of each algorithm, and the color of the circle represents the predicted class by the algorithm.

\end{document}